\documentclass[conference]{IEEEtran}
\usepackage[top=0.75in, bottom=1.05in, left=0.625in, right=0.625in]{geometry}
\IEEEoverridecommandlockouts
\usepackage{cite}
\usepackage{xcolor,soul,framed}
\usepackage{listings}
\colorlet{shadecolor}{yellow}
\usepackage[pdftex]{graphicx}
\graphicspath{{../pdf/}{../jpeg/}}
\DeclareGraphicsExtensions{.pdf,.jpeg,.png}
\usepackage{amssymb}
\usepackage[cmex10]{amsmath}
\usepackage{array}
\usepackage{mdwmath}
\usepackage{mdwtab}
\usepackage{eqparbox}
\usepackage{url}
\usepackage{algorithm}
\usepackage{algorithmic}
\usepackage{svg}

\begin{document}
\bstctlcite{IEEEexample:BSTcontrol}
    \title{Bridge to Real Environment with Hardware-in-the-loop for Wireless Artificial Intelligence Paradigms}
\vspace{0.1cm}

\author{Jeffrey Redondo\IEEEauthorrefmark{1} (Student Member, IEEE), Nauman Aslam\IEEEauthorrefmark{1} (Senior Member, IEEE),\\
Juan Zhang\IEEEauthorrefmark{1} (Member, IEEE), Zhenhui Yuan\IEEEauthorrefmark{2} (Member, IEEE), David Jimenez\IEEEauthorrefmark{3} \\
\IEEEauthorblockA{\IEEEauthorrefmark{1} University of Northumbria, Newcastle, UK, \IEEEauthorrefmark{2} University of Warwick, Coventry, UK}, 
\IEEEauthorblockA{\IEEEauthorrefmark{3} V-TRON Company, Netherlands}
Email: \texttt{\{jeffrey.redondo, nauman.aslam, juan.zhang\}@northumbria.ac.uk}, \\
\texttt{Zhenhui.Yuan@warwick.ac.uk}, \texttt{d.jimenez@v-tron.eu}}
  
\maketitle

\fbox{\parbox{\dimexpr\columnwidth-10\fboxsep-0\fboxrule\relax}{
    This work has been submitted to the IEEE for possible publication. Copyright may be transferred without notice, after which this version may no longer be accessible.
  }}
\\
\\
\\
\begin{abstract}
Nowadays, many machine learning (ML) solutions to improve the wireless standard IEEE802.11p for Vehicular Adhoc Network (VANET) are commonly evaluated in the simulated world. At the same time, this approach could be cost-effective compared to real-world testing due to the high cost of vehicles. There is a risk of unexpected outcomes when these solutions are implemented in the real world, potentially leading to wasted resources. To mitigate this challenge, the hardware-in-the-loop is the way to move forward as it enables the opportunity to test in the real world and simulated worlds together. Therefore, we have developed what we believe is the pioneering hardware-in-the-loop for testing artificial intelligence, multiple services, and HD map data (LiDAR), in both simulated and real-world settings.
\end{abstract}


\IEEEpeerreviewmaketitle

\section{Introduction} \label{intro}
Researchers have been dedicated to developing new machine learning algorithms to provide the best Quality of Service (QoS) for VANET. This is because, in the near future, autonomous vehicles (AVs) will be on the road \cite{gov_uk} in many developed countries worldwide. However, it is imminent that the increase in AVs creates a dense network, which poses significant challenges due to the substantial costs involved in conducting tests using real vehicles, and wireless devices. To mitigate this challenge, researchers have been merging the simulated work and real work by developing a hardware-in-the-loop testbed including real vehicles for testing hardware \cite{SDF_emulation,emulator_edge_computing,emulatio_eco_driving} or wireless network for the dissemination of data in a VANET. This approach aims to yield invaluable insights into solution behavior in real-world scenarios, ultimately constraining the expenses associated with real network trials. 
One of the primary obstacles faced by VANET is the fixed contention window (CW) size which increases the packet collision as studied in \cite{ieee80211_cw2}. To address this, several investigations have concentrated on finding a suitable CW value to be assigned in a dense environment such as \cite {low_latency_new_ac,adaptive_edca,q_learning_fairness}. However, their implementation incurs in changing or modifying the existing standard. Integrating these solutions into the real world demands extensive testing in a real environment before deployment, which could be costly as a lot of devices must be bought or built. Furthermore, if testing uncovers inefficiencies, the associated losses could be substantial. 
To tackle these challenges, we are developing a testbed that allows the integration of real devices and scenarios into simulated environments. This approach is poised to reduce the time and costs involved in bringing solutions or products to the market. This paper provides the resources to test any solution such as multiple access \cite{TDMA}, new access category \cite{low_latency_new_ac}, game theory solution \cite{adaptive_edca}, and machine learning algorithms such as Q-Learning \cite{q_learning_fairness, our_sojourn_single_agent}, etc within the simulated and real environment by using hardware-in-the-loop (HIL) technique.

\subsection{Contributions} 
\begin{itemize}
\item We developed one of the first hardware-in-the-loop testbed for dissemination of LiDAR data to test artificial intelligent algorithms in a VANET using OMNet++, python, UDP server/client, and Robotic Operation Systems (ROS).
\item We demonstrated the functionality of the testbed on the dissemination of two different services LiDAR data, and video data. The evaluation is shown in terms of delay and throughput.
\end{itemize}

\section{Related Work} \label{related_work}
In this section, it is described several work related to simulations and hardware-in-the-loop for wireless network.

\subsection{Wireless Network Testbed}
There have been several efforts to improve the testing of how new solutions could impact the real world before their integration into large-scale environments such as VANET. Decades ago, these efforts involved the implementation of  IEEE802.11p in an embedded system \cite{embedded_system_emulation}, and software-defined radio (SDF) \cite{SDF_emulation}. These approaches leveraged real hardware and network setups in conjunction with software to avoid the use of extra hardware or to reduce it, aiming to reduce market costs and streamline testing processes. Consequently, authors have been incorporating the real and simulation worlds to continually reduce market costs and circumvent the need for extensive and costly equipment to evaluate new solutions. Notably, in a recent study \cite{emulator_edge_computing} authors have included the traffic simulator SUMO \cite{SUMO} to facilitate a more realistic practical test of an edge computing solution which is in the physical world, known as the Cloud-in-the-Loop (CiL). According to the authors, this method enables the evaluation of the system before testing with real vehicles. Furthermore, emerging research such as \cite{emulatio_eco_driving} has demonstrated the benefits of having a HIL testbed by testing a device called Cohda Wireless MK5 in conjunction with a cooperative eco-driving application in an emulated scenario.

These studies have demonstrated advancement in integrating HIL to test real applications in a simulated environment. Nevertheless, none of them have focused on testing and evaluating novel machine learning algorithms designed to mitigate delay increase and throughput decrease. To address this challenge, we have developed a testbed using HIL that can incorporate real devices, such as LiDAR data, and edge computing.

\section{Overview} \label{overview}
In this section, an overview of reinforcement learning is introduced.

\subsection{Reinforcement Learning}
Reinforcement learning is a paradigm of Machine learning that involves the use of states, actions, and reward function for the learning purpose of an agent. A tuple $(S,A,P,R)$ is defined, $S$ stands for the set state, $A(s)$ is the set of actions which is dictated by the policy $P_{\pi}[A = a|S=s]=\pi(a|s)$. $P(s,s')$ is the transition probability to move to the next state $s'$ based on the selected action $a \in A(s)$. Finally, $R_{s}^{a}(s,s')$ refers to the reward obtain from moving from $s$ to $s'$ with action $a$ \cite{q_learning_fairness}. 

Given the stochastic nature of the vehicular network, the $P(s,s')$ for a VANET is unknown due to the high dynamic of the traffic flow of the vehicles. Thus, it is imperative to determine the optimal policy $\pi$. In \cite{our_sojourn_single_agent}, we opted for off-policy TD control Q-learning, which directs the agent to seek out the optimal policy $\pi$.

\begin{figure}[h]
  \begin{center}
  \includegraphics[width=3in]{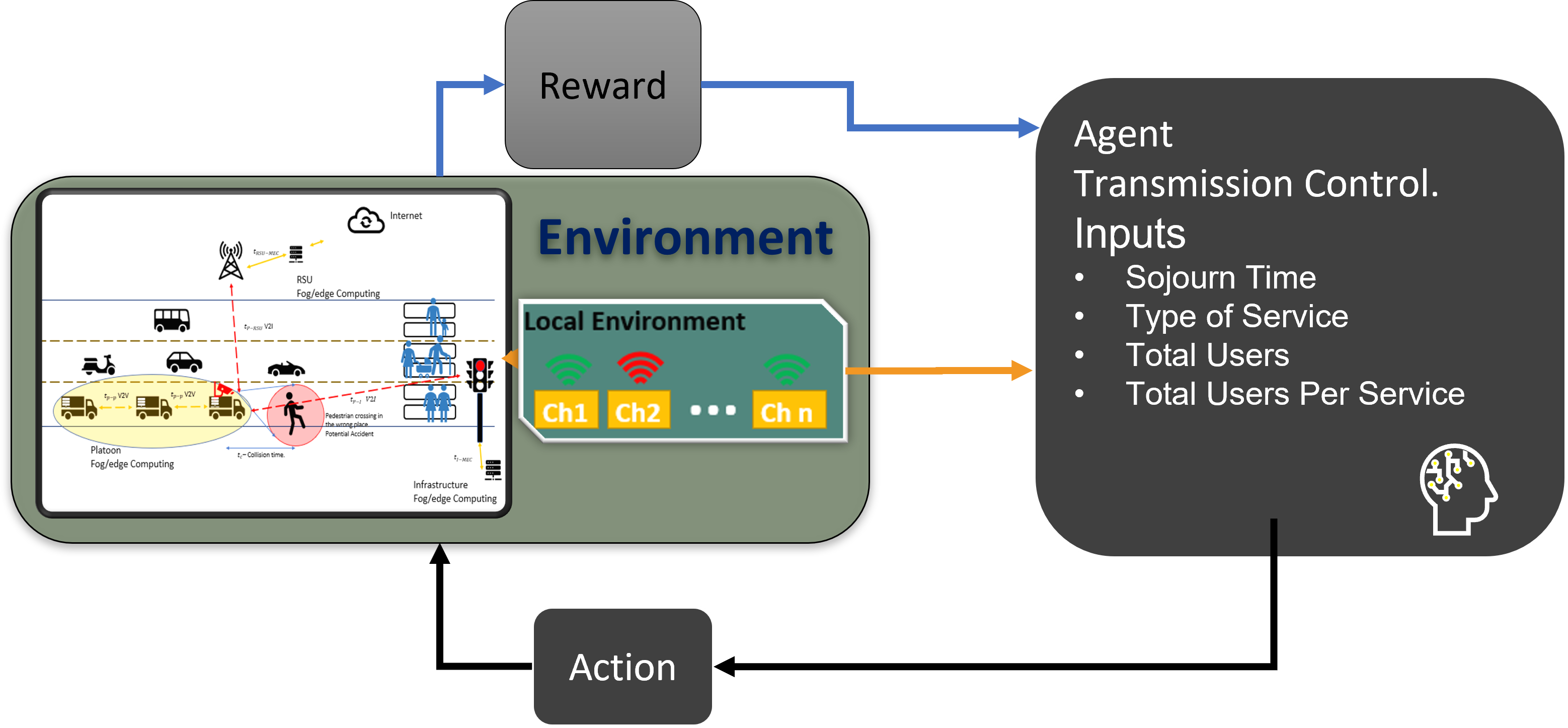}
  % \vspace{5pt}
  \caption{Reinforcement Learning Architecture.}
  \label{fig:reinforcemet_learning_architecture}
  \end{center}
\end{figure}

In Fig. \ref{fig:reinforcemet_learning_architecture} is displayed the ML architecture for RL. The Q value table is updated using equation \ref{eq_q_learning}, that contains the value of the next state with a discount factor $\gamma$, and a learning rate $\alpha$. 
\vspace{0.01cm}
\begin{equation}
    Q(s,a) = Q(s,a) + \alpha\left[r+\gamma*max_{a'}Q(s',a')-Q(s,a)\right]
\label{eq_q_learning}
\end{equation}
\vspace{0.01cm}
The discount factor dictates the importance of future rewards, and the learning rate determines how fast the $Q$ values is updated.

\section{Design} \label{design}
In this paper, we designed and implemented a testbed that involves the use of physical hardware to transfer data to the virtual environment. To accomplish this, we utilized OMNet++ which allows us to use an external network interface via virtual interfaces in a Linux environment. With this approach, we can test a real device within a bigger wireless network that could be manipulated in the simulation, for example, we could vary the number of vehicles. Furthermore, to test an application developed to improve the network performance, we have selected our previous paper \cite{our_sojourn_single_agent} that suggests a QoS solution that operates at the application layer without demanding any MAC layer modifications of the IEEE802.11p standard.

\subsection{Protocol Layer Structure \& IP Network Desing}
\begin{figure}[h]
\centering
\includegraphics[width=\linewidth]{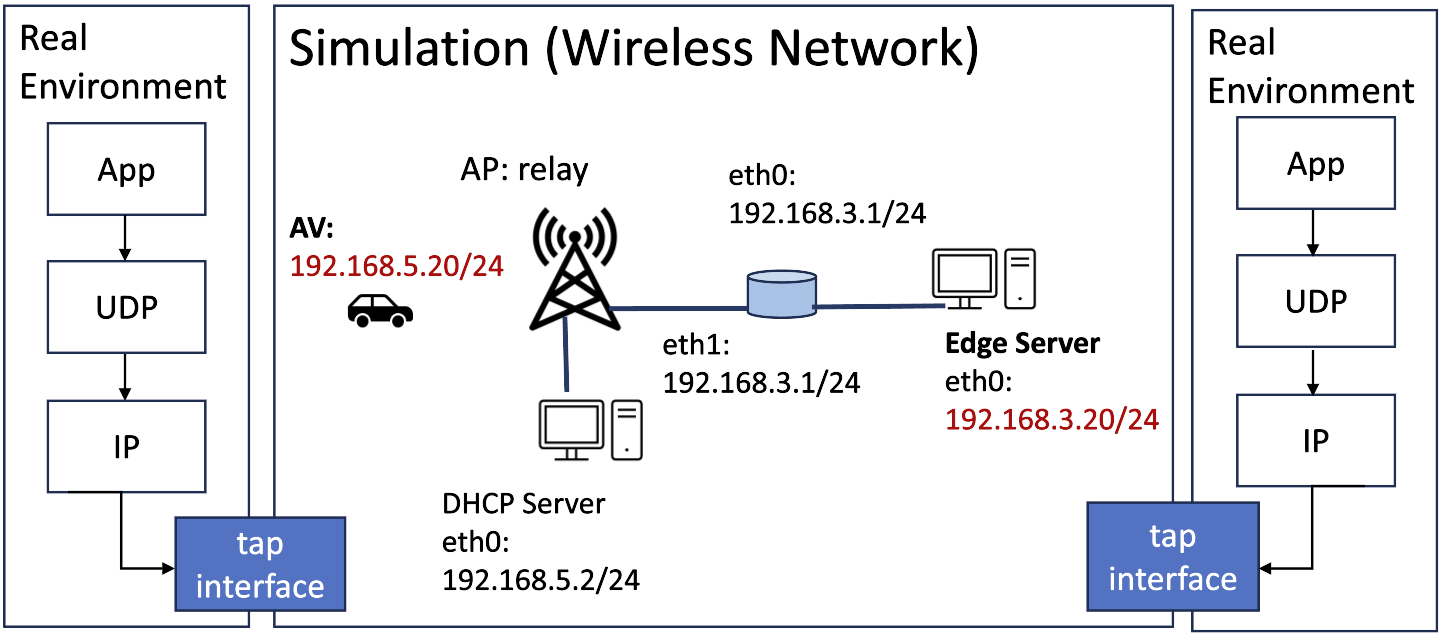}%
\caption{Network architecture of the real environment and simulated one.}
\label{fig:network_diagram}
\end{figure}

To accomplish the development of the testbed, we have designed the network configuration as described in Fig. \ref{fig:network_diagram}. In this network diagram illustrates the interconnection of the real and simulated environment through the TAP interface. We have selected the TAP interface due to its characteristic to emulate a layer 2 device and to interconnect virtual machine to the external environment.
\vspace{0.01cm}
\vspace{0.01cm}
The IPs highlighted in red belong to the devices that operate in the real environment through the TAP interfaces, which also should be configured in OMNet++ through configuration Extensible Markup Language (XML) file or Dynamic Host Configuration Protocol (DHCP) server. In our case, we utilize both DHCP for the other devices in the network and configuration XML file for the specific IP assignment. As illustrated in Fig. \ref{fig:network_diagram}, the AV is attached to the TAP interface, while the edge server to the other. This particular setup allows us to assess the impact a vehicular network could have while a real application is tested in a simulated VANET scenario. As a result, it provides insights into the performance of new algorithms in a more realistic environment before their deployment in the real world.
 
\section{Setup \& Simulation} \label{simulation}
To integrate the real and simulated environment, the external interfaces provided by the OMNet++ version 6 \cite{omnetppOMNeTDiscrete} and Inet Framework version 4 \cite{omnetppINETFramework} were utilized. It provides the physical and MAC layer stack for IEEE802.11p. To evaluate the reinforcement learning algorithm in \cite{our_sojourn_single_agent}, we have utilized the same tool to integrate Python and OMNet++ developed by authors in \cite{veinsgym}. Please refer to Table \ref{table:simulation_parameters} for a detailed breakdown of our simulation parameters. 

\subsection{Linux \& OMNet++ External Interface Configuration}
\begin{figure}[h]
\centering
\includegraphics[width=\linewidth]{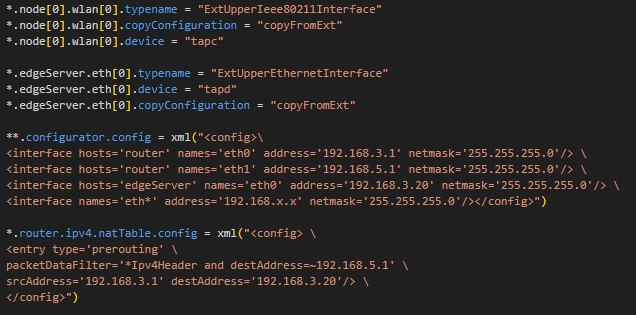}%
\caption{Configuration for the external interface, static IPs, and router forwading rule.}
\label{fig:interface_configuration}
\end{figure}

For the involvement of the external devices in the simulation, we have configured the emulation example provided within the inet library. For the edge serve, the external module is \textbf{ExtUpperEthernetInterface}, and for the AV wireless interface is \textbf{ExtUpperIeee80211Interface}. The configuration is displayed in Fig. \ref{fig:interface_configuration} together with the configuration XML file for the static IPs, and router forwarding rule. Additionally, the \textbf{scheduler-class = "inet::RealTimeScheduler"}, should be added to the ini file to work properly.
This configuration requires the setup of virtual interfaces in Linux called TAP which emulates layer 2 devices. In this case, OMNet++, inet library allows us to connect to the real world. The configuration for our setup is the following,
\vspace{0.01cm}
\begin{lstlisting}[language=bash]

# create TAP interfaces
sudo ip tuntap add mode tap dev tapc
sudo ip tuntap add mode tap dev tapd

# Assign IP addresses to interfaces
sudo ip addr add 192.168.5.20/24 dev tapc
sudo ip addr add 192.168.3.20/24 dev tapd

# Bring up all interfaces
sudo ip link set dev tapc up
sudo ip link set dev tapd up
\end{lstlisting}
\vspace{0.01cm}
The commands are for Linux terminal. For more details in the devices connections, the interfaces are connected as described in Fig. \ref{fig:network_diagram}, and \ref{fig:interface_configuration}. To properly evaluate the hardware-in-the-loop and the reinforcement learning, we utilized iperf, a tool for measuring network performance, in the external device configured as a server and configured as a client at the receiver. It also facilitates the learning of the ML model as we could simulate any application by changing the bandwidth value. The commands are the following,
\vspace{0.01cm}
\begin{lstlisting}[language=bash]
# Client configuration
iperf -c 192.168.5.1 -B 192.168.5.20 
    -u -b 22M -t 200
# Server Configuration
iperf -s -e -i 1 -u -B 192.168.3.20
\end{lstlisting}
\vspace{0.01cm}
The analysis of the simulation is performed in terms of delay, and throughput Key Performance Indicators (KPIs). delay is computed by subtracting the difference between the generated packet time, and the arrival time at the destination.

\begin{table}[ht]
% \begin{table}[htbp]
\caption{Simulation Parameters \cite{our_sojourn_single_agent}}
\begin{center}
\begin{tabular}{|c|c|}
\hline
\textbf{\textit{Parameter}} & \textbf{\textit{Value}} \\
\hline
Tile Dimension & 300x100 meters  \\
\hline
Episodes & 3  \\
\hline
$\epsilon-greedy$ & 0.2  \\
\hline
Discount Factor $\gamma $& 0.99  \\
\hline
Learning Rate $\epsilon $& 0.1  \\
\hline
$\alpha_1 $, $\alpha_2 $ & 0.3 and 0.7 respectively  \\
\hline
Simulation time & 250s  \\
\hline
Vehicle Density$^a$ & varies according traffic flow  \\
\hline
Coverage Area & 200m  \\
\hline
Vehicle max speed & 17m/s  \\
\hline
Vehicle Acceleration  & {2.6 m}/{$s^2$} \cite{acceleration_deceleration_petrol_car} \\
\hline
Vehicle Deceleration  & {4.5 m}/{$s^2$} \cite{acceleration_deceleration_petrol_car}  \\
\hline
Tx Power & 200 mW  \\
\hline
Frequency & 5.9 GHz  \\
\hline
Bandwidth & 20MHz  \\
\hline
Best-effort data rate & 28Mbps  \\
\hline
data rate & 1.37Mbps \cite{5gaa_delay}\\
\hline
TXOP limit & Disabled as per standard  \\
\hline
Number of hidden layers & 1 \\
\hline
Number of neurons in hidden layer & 32 \\
\hline
Buffer size & 500  \\
\hline
%\multicolumn{2}{l}{$^{\mathrm{a}}$Total number of vehicles through the simulation was 165.} \\
% \multicolumn{2}{l}{$^{\mathrm{b}}$Average value of petrol cars\cite{acceleration_deceleration_petrol_car}.}
\end{tabular}
\label{table:simulation_parameters}
\end{center}
\end{table}

\subsection{Scenarios for Validation}
We have selected different scenarios to test the hardware-in-the-loop testbed in a VANET with HD Map updates. To this end, we opted for testing one vehicle as bench-marking, and then we increased the number of AVs to 2, 3, 5, 7, and 10. The selected number of vehicles was sufficient to illustrate how the network performance is affected by the increase of nodes in the network. Besides, it is also limited by the computer capacity used for the simulation. For the simulation, we have assigned the same number of bytes to be transmitted for each vehicle at the same interval time. The details of the simulation parameters are in Table \ref{table:simulation_parameters}.
\vspace{0.1cm}
\subsubsection{One vehicle}
In this scenario, we corroborate that the setup was working as designed, and to provide a benchmark for the comparison when the vehicular density increases. 
\vspace{0.1cm}
\subsubsection{Variable Vehicular Density}
The second scenario involves an increase in the number of vehicles. For this scenario, one vehicle is connected to the external device, and the others are configured to transmit one particular packet size. It is paramount to note that we have not enabled the EDCA.
\vspace{0.1cm}
\subsubsection{RL learning algorithm Evaluation}
After having all the data from the previous setup, we evaluate the impact of a machine learning solution studied in \cite{our_sojourn_single_agent}. The algorithm is modified to only 4 actions for simplicity in this setup. This scenario provides the validation that our HIL setup could be utilized to test any solution in a more realistic environment.
\vspace{0.1cm}
\subsubsection{RL learning algorithm with LiDAR}
As we have developed the tool for HD Map updates, we have implemented a small AV to gather LiDAR data and transfer it to the edge server via Ethernet. This allows the data to be accessed from a PC and shared with the simulator using the TAP interface. Additionally, we utilized Robot Operating System (ROS2) Robot Vizualization (RVIZ) for visualizing the LiDAR data.
\vspace{0.1cm}
\subsubsection{RL learning algorithm with video}
Additionally, we also conducted a test case involving the real-time streaming of a video in order to explore an alternative application of the HIL setup.

\section{Result and Discussion} \label{results}
In this section, the results are illustrated for the different scenarios. Additionally, the novel Q-learning algorithm developed in \cite{our_sojourn_single_agent} for reducing delay has been tested, which does not need modification for the wireless technology IEEE802.11.

\subsection{Validation Scenarios} 
There are three validation setups described in the previous section that involve varying the number of vehicles, and how this affects the different applications. These applications are iperf, LiDAR, and video.

\subsubsection{Validation vehicular density and RL algorithm} 
Firstly, the impact the number of vehicles has over the network performance was evaluated with the commonly used application iperf. Thus, the number of vehicles varies from 2 to 10. It is paramount to mention that in all the simulations, the vehicle "node[0]" (in OMNet++ ini file) is the one connected to the external interface representing an external AV. The bandwidth configuration for iperf client was set to \textbf{22M}. 

\begin{figure}[h]
\centering
\includegraphics[width=\linewidth]{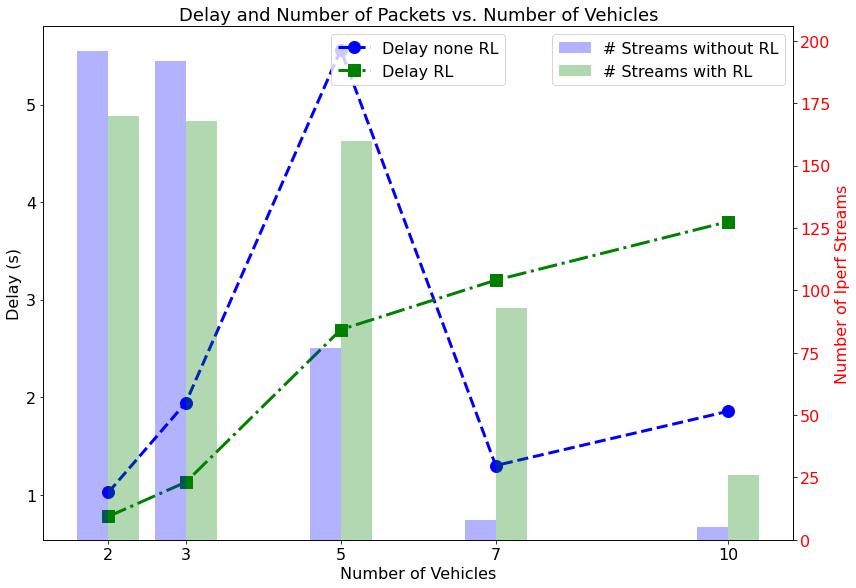}%
\caption{Delay and packet streams results using iperf tool.}
\label{fig:delay}
\end{figure}

\paragraph{Delay}
As it is illustrated in Fig. \ref{fig:delay}, the delay increments while the number of vehicles increases, while the number of received streams reduces. Nevertheless,It's worth noting that the delay was lower than expected when there were 10 AVs in motion, compared to the scenario where no RL algorithm was used. This is because the number of streams received by the edge server is less than the one received by the previous configuration with fewer vehicles. This indicates the wireless node was granted access to the network achieving lower delay. On the other hand, for 7 AVs the delay was high, suggesting more network collisions despite access to the network.

In comparison, with the results using the modified RL algorithm with 4 actions only, an improvement is observed in terms of delay for most of all the vehicular densities.The RL solution produces a decrease in delay of $24.1\%$, $41.9\%$, $51.58\%$, for $2$, $3$, and $5$ vehicles respectively. However, there is an exception with the vehicular density of 7, and 10. The number of streams received is lower compared to the other vehicular densities. If we analyze the increase in the number of received streams, the external device was able to access the medium and transmit a higher number of streams while the RL was functioning. The percentage difference in the number of streams is $70\%$, $168\%$, and $135\%$ for 5, 7, and 10 vehicles respectively.

\begin{figure}[h]
\centering
\includegraphics[width=\linewidth]{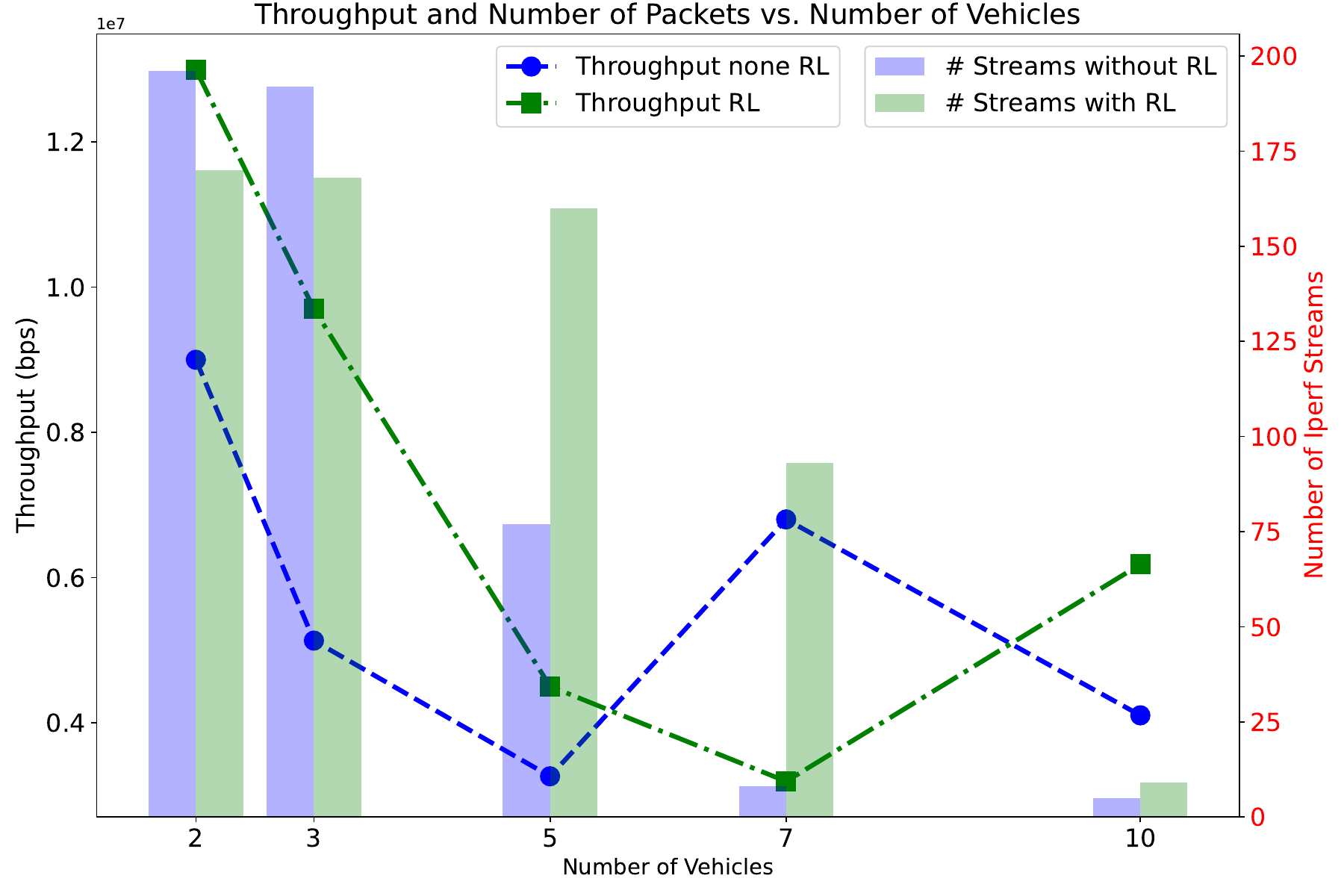}%
\caption{Throughput and packet streams results using iperf tool.}
\label{fig:throughput}
\end{figure}

\paragraph{Throughput}
We can observe in Fig. \ref{fig:throughput}, that the throughput is higher in almost all the vehicular density, except 7 and 10 vehicles. This is related to the number of received streams. As mentioned the increase in the number of streams was higher for these vehicular densities. Therefore, the average throughput is affected by having low performance during some parts of the simulation. It is also paramount to mention that the majority of the received stream arrived successfully on the server when the simulation started, after that, the other vehicles occupied the network to more extent.

For the vehicular density of $2$, $3$, and $5$ the throughput increases by $44.44\%$, $89.08\%$, $38.04\%$ respectively. However, at a density of $7$, and $10$, the throughput decreases by $28\%$, and $7.8\%$ respectively, compared to the normal setup. This increase is in line with the previous delay analysis, where the number of packets increases.

\subsubsection{Validation LiDAR data} 
\begin{figure}[h]
\centering
\includegraphics[width=\linewidth]{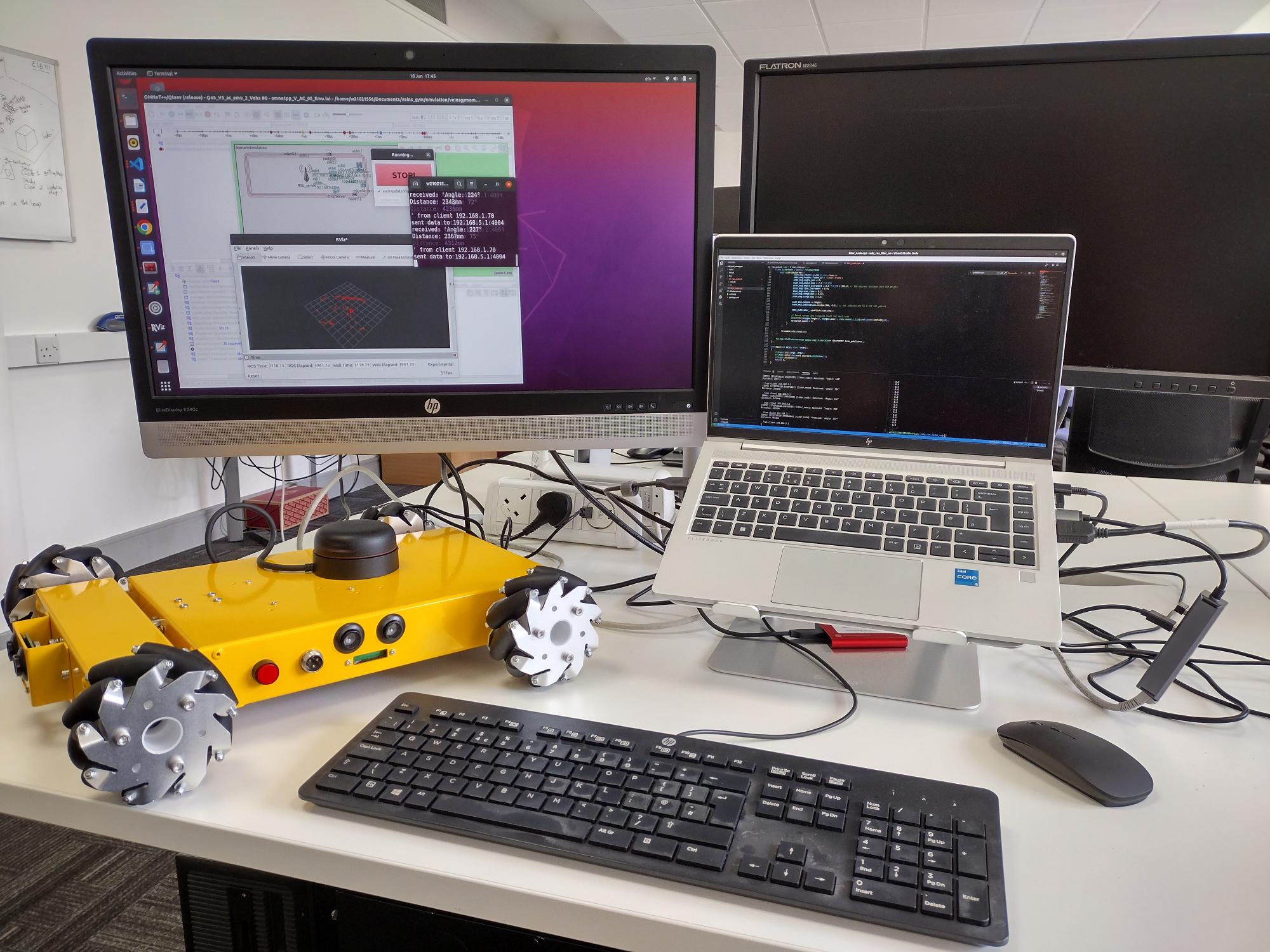}%
\caption{Real and simulated setup with AV, LiDAR and PCs.}
\label{fig:lidar_setup}
\end{figure}

After evaluating the HIL with iperf, we proceeded to test a new application that transfers LiDAR data through the wireless network by the AV. This application is an essential component in realizing a higher level of automation for AVs \cite{nvidea_hdmap}. Consequently, we decided to conduct tests to quantify the impact of utilizing an RL solution, in this case, the solution in \cite{our_sojourn_single_agent}, on the number of messages received. Fig. \ref{fig:lidar_setup} displays the setup where an AV gathers the LiDAR data and transfers it via User Datagram Protocol (UDP). In our case to test the emulator testbed, we use the Ethernet port of the micro-controller board. The data is received and sent through the OMNet++ simulation using the TAP configuration and external modules. When the data reaches the destination, the information is received by a UDP server, which transfers the data to the ROS2 RVIZ \cite{rviz_ros} laser scanner data visualizer. For the micro-controller board, we used ARM with the collaboration of the company V-TRON \cite{v_tron}, who provided the board and helped us developing the code to gather and transfer the data from LiDAR.

\begin{figure}[h]
\centering
\includegraphics[width=\linewidth]{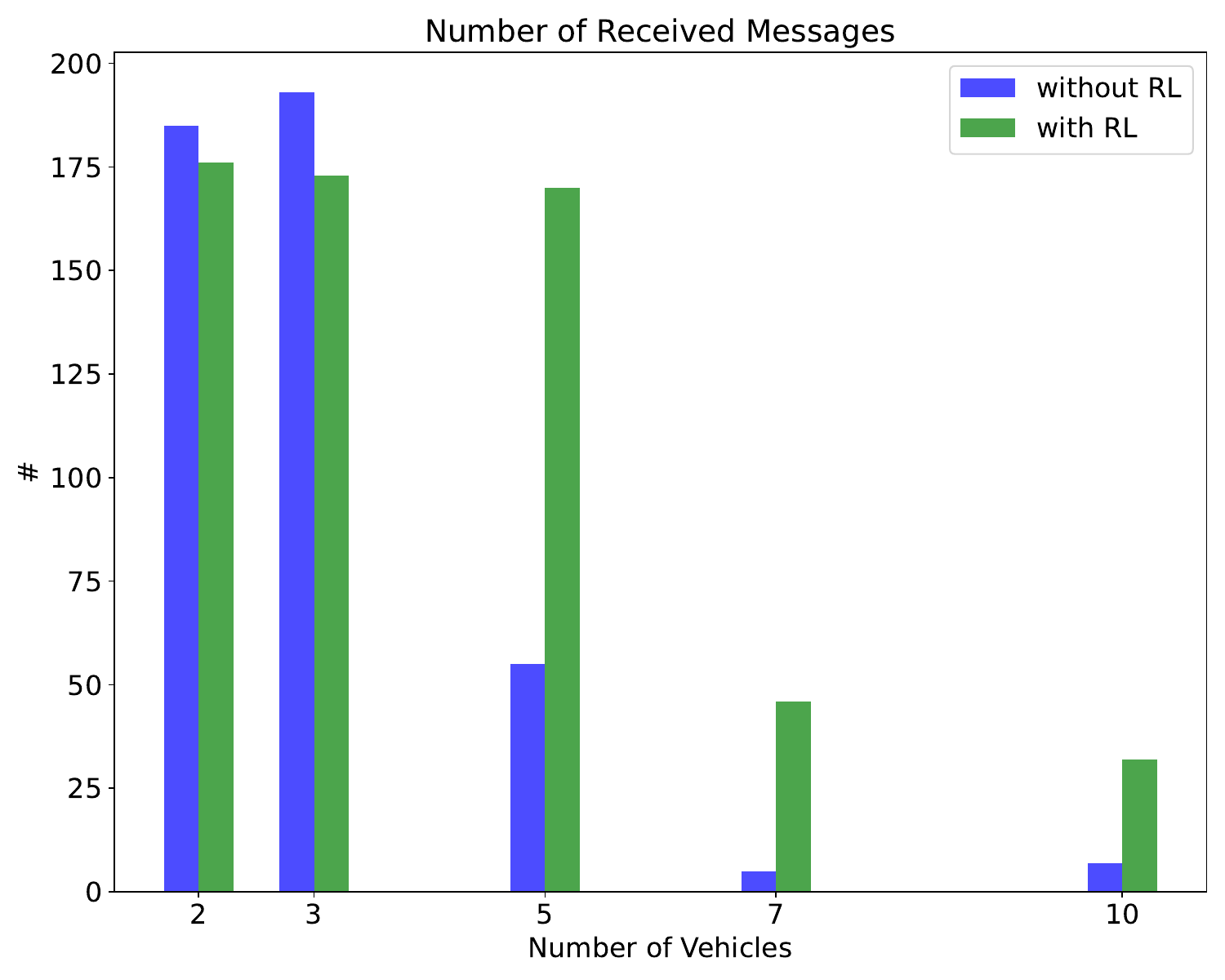}%
\caption{Number of packets received using LiDAR data and RVIZ ROS visualizer tool.}
\label{fig:lidar_stats}
\end{figure}

The results for the LiDAR application validation are displayed in Fig. \ref{fig:lidar_stats}. It is clearly observed how the RL algorithm could contribute to higher delivery of messages to the RVIZ application. The increase for a vehicular density of $2$ was $38\%$, for $3$ was $49\%$, for $5$ was $135\%$, for $7$ was $46\%$, and for $10$ was $50\%$. This results reveals the effectiveness of the AI algorithm and the usefulness of the HIL testbed.

\subsubsection{Validation with another type of simulation} 
As we know, video streaming is an important application nowadays, especially for AVs for constructing maps. For this setup, we utilized the VLC video software to transmit video through OMNet++, and also to receive the streamed video. The command for this could be found in the inet framework example called video streaming. You can refer to Figure 1 for a visual representation of vehicles in motion in OMNet++ and the received video. To assess the performance, we evaluated to compare the amount of data received at the server with and without the reinforcement learning (RL) algorithm for comparison.

\begin{figure}[h]
\centering
\includegraphics[width=8cm, height=6cm]{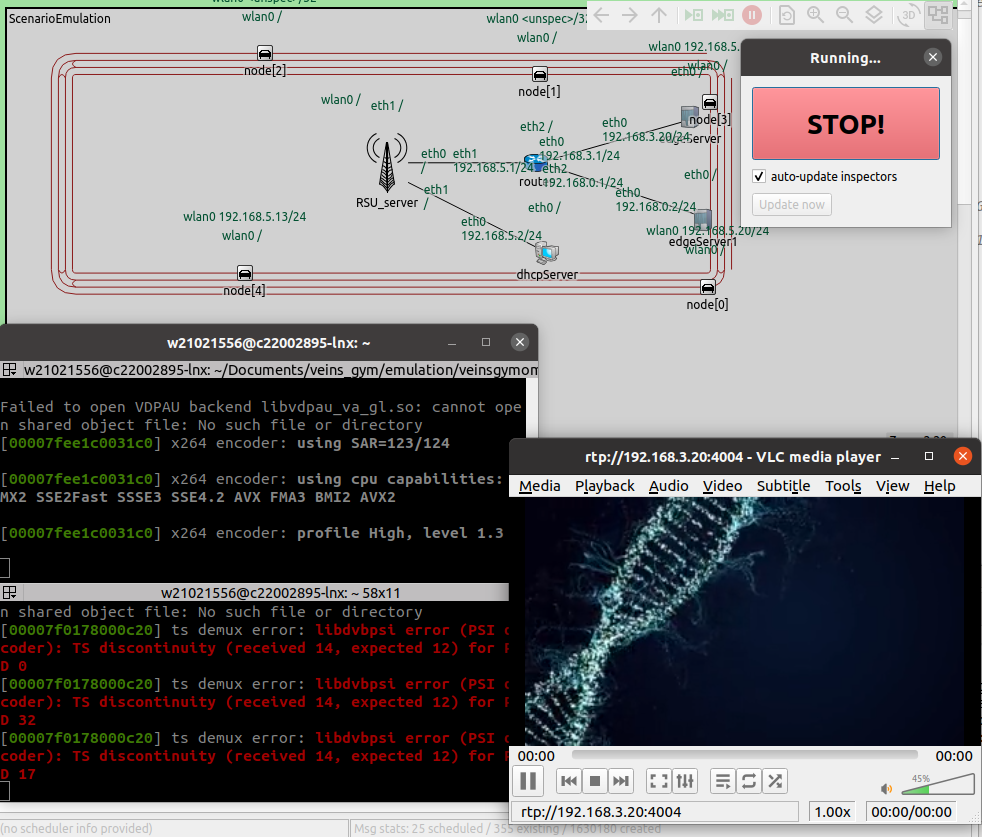}%
\caption{Architecture of the real environment and simulated one.}
\label{fig:video_simu}
\end{figure}

\begin{figure}[h]
\centering
\includegraphics[width=\linewidth]{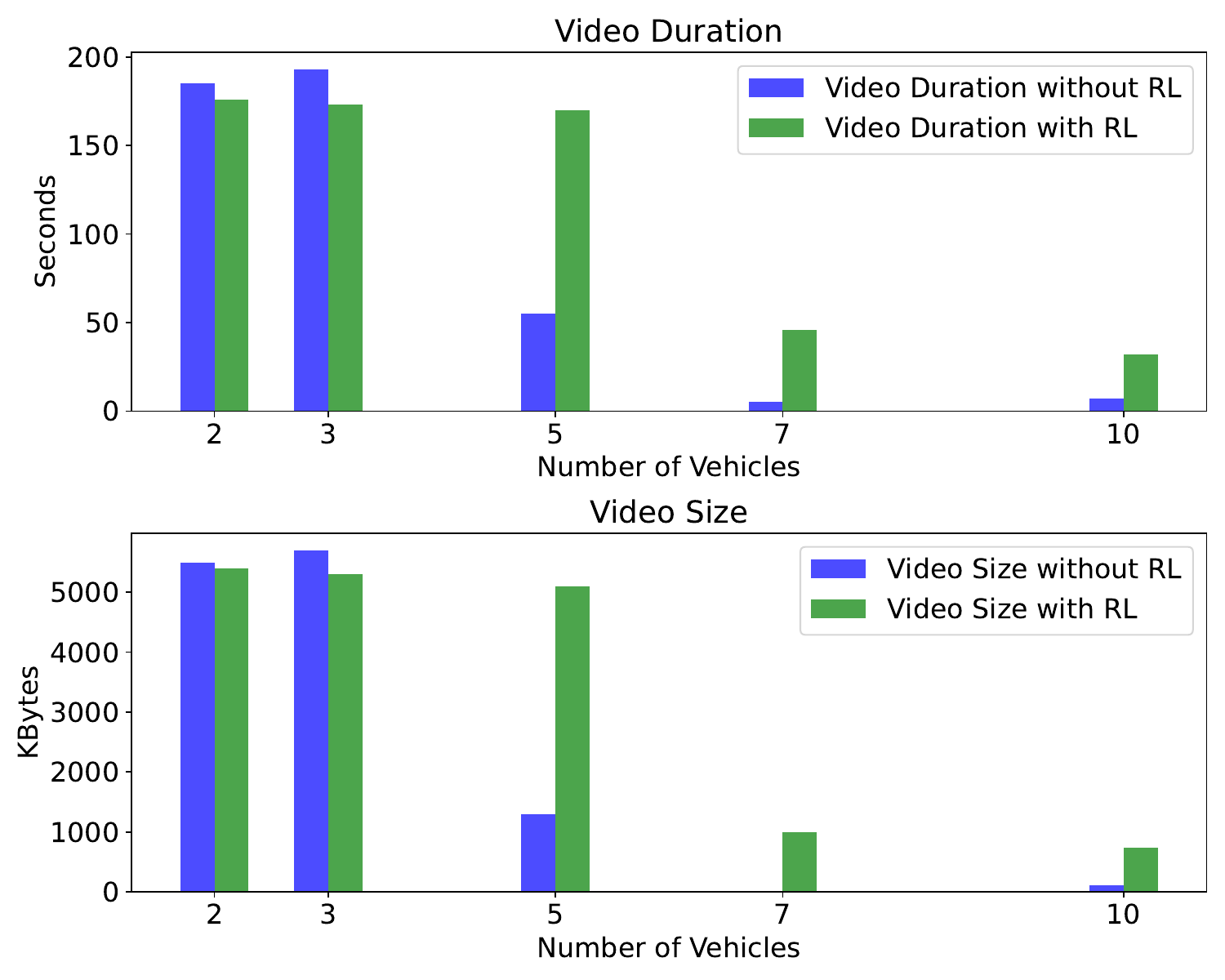}%
\caption{Duration of the video and number of bytes received while streaming a video.}
\label{fig:video_stats}
\end{figure}

The results are illustrated in Fig. \ref{fig:video_stats}.  Upon implementing the RL algorithm, enhancements in bytes received and video duration were evident. The highest improvements are seen in the vehicular density of $5$, $7$, and $10$. For $2$, and $3$ vehicles, there is a decrease in the duration of the video corresponding directly to the size of the video, this decrease corresponds to approximately $4.8\%$ for $2$, and $10\%$ for $3$. In terms of the video size, the decrease is about $1.8\%$ for $2$ vehicles and $7\%$ for $3$ vehicles. On the contrary, for $5$, $7$, and $10$ vehicles, the percentage differences were notably higher, around $169\%$, $176\%$, and $145\%$ respectively.

\section{Conclusion} \label{conclusion}
Our research has led to the successful development of a cutting-edge HIL testbed, integrating both real-world and simulated components. The results conclusively demonstrated that the simulated RL algorithm significantly enhanced network performance when applied to real-world scenarios. Our next step in our investigation is to incorporate a real network into the simulation, further validating the robustness and efficacy of our approach.

\section*{Acknowledgment}
This work was partly funded by EPSRC with RC Grant reference EP/Y028023/1, UKRI under grant number EP/Y028023/1, the European Horizon2020 MSCA programme under grant agreement  No. 101086228. Collaboration with V-TRON company.

\ifCLASSOPTIONcaptionsoff
  \newpage
\fi

\bibliographystyle{IEEEtran}
\bibliography{IEEEabrv,Bibliography}

\end{document}